\documentclass{article}

\usepackage{arxiv}

\usepackage[utf8]{inputenc} 
\usepackage[T1]{fontenc}    
\usepackage{hyperref}       
\usepackage{url}            
\usepackage{booktabs}       
\usepackage{amsfonts}       
\usepackage{nicefrac}       
\usepackage{microtype}      

\usepackage{graphicx}
\usepackage{natbib}
\usepackage{doi}
\usepackage{caption}
\usepackage{algorithm}
\usepackage{amsmath}
\usepackage{mathtools}
\usepackage[noend]{algpseudocode}
\usepackage[dvipsnames,svgnames]{xcolor}
\definecolor{aisle}{HTML}{FFFFE5}
\definecolor{door}{HTML}{F7FCB9}
\definecolor{storage}{HTML}{004529}
\DeclareRobustCommand{\legendsquare}[1]{%
  \textcolor{#1}{\rule{1ex}{1ex}}%
}
\algnewcommand{\Inputs}[1]{%
  \State \textbf{Inputs:}
  \parbox[t]{.8\linewidth}{\raggedright #1}
}
\algnewcommand{\Initialize}[1]{%
  \State \textbf{Initialize:}
  \parbox[t]{.8\linewidth}{\raggedright #1}
}

\usepackage{xcolor}

\title{A Novel Framework for Automated Warehouse Layout Generation}

\author{
    Atefeh Shahroudnejad\thanks{Corresponding author}\\
    Amii (Alberta Machine Intelligence Institute)\\
    \texttt{atefeh.shahroudnejd@amii.ca}\\
    \And
    Payam Mousavi \\
    Amii (Alberta Machine Intelligence Institute)\\
    \texttt{payam.mousavi@amii.ca}\\
    \And
    Oleksii Perepelytsia\\
    Routeique Inc\\
    \texttt{perepelytsia.al@gmail.com}\\
    \And
    Sahir\\
    Amii (Alberta Machine Intelligence Institute)\\
    \texttt{sahir@amii.ca}\\
    \And
    David Staszak\\
    Amii (Alberta Machine Intelligence Institute)\\
    \texttt{david.staszak@amii.ca}\\
    \And
    Matthew E.~Taylor\\
    Amii (Alberta Machine Intelligence Institute)\\
    University of Alberta\\
    \texttt{matthew.e.taylor@ualberta.ca}\\
    \And
    Brent Bawel
    Routeique Inc\\
    \texttt{brent.bawel@routeique.com}\\
}

\date{}

\hypersetup{
pdftitle={A Novel Framework for Automated Warehouse Layout Generation},
pdfauthor={Atefeh Shahroudnejad, Payam Mousavi, Oleksii Perepelytsia, Sahir, David Staszak, Matthew E.~Taylor, Brent Bawel},
pdfkeywords={AI, constrained optimization, automation, warehouse design, logistics},
}

\begin{document}
\maketitle

\begin{abstract}
	Optimizing warehouse layouts is crucial due to its significant impact on efficiency and productivity. We present an AI-driven framework for automated warehouse layout generation. This framework employs constrained beam search to derive optimal layouts within given spatial parameters, adhering to all functional requirements. The feasibility of the generated layouts is verified based on criteria such as item accessibility, required minimum clearances, and aisle connectivity. A scoring function is then used to evaluate the feasible layouts considering the number of storage locations, access points, and accessibility costs. We demonstrate our method's ability to produce feasible, optimal layouts for a variety of warehouse dimensions and shapes, diverse door placements, and interconnections. This approach, currently being prepared for deployment, will enable human designers to rapidly explore and confirm options, facilitating the selection of the most appropriate layout for their use-case.
\end{abstract}


\section{Introduction}
The main goal of Warehouse Management Systems (WMS) is running operations as efficiently as possible to improve profitability through increasing productivity, reducing labor costs, and ultimately increasing customer satisfaction.
One of the key components of a WMS is optimal space utilization. It reduces the need for a larger capacity warehouse by maximizing inventory storage and minimizing wasted or underutilized areas. Moreover, warehouse configuration has a direct impact on all warehouse operations, especially the worker routing and picking processes. Efficient warehouse configurations (i.e., layouts) can enhance the order fulfillment process by eliminating unnecessary movement and related errors, resulting in time and cost savings \cite{mohamud2023role,richards2017warehouse}. 
However, the vast majority of warehouses worldwide still continue to rely on manual management or basic automation ~\cite{albert2023trends}.  There are a range of traditional and non-traditional manual layout designs that have been proposed over the years to speed up warehouse operations and minimize operating costs~\cite{gue2009aisle, bortolini2020integration, kocaman2021aisle, zhang2021designing}. Although manually-designed layouts might be feasible for small warehouses with limited options, for larger scales, they are a less efficient use of warehouse designer time and more prone to human error. Hence, an automated process for candidate layout generation would be beneficial to all stakeholders.
Moreover, an automated warehouse layout process would allow users to change specifications over time to meet shifting product demand and facilitate the expansion or restructuring of the warehouse while preserving efficiency and productivity.

To automatically generate optimal candidate layouts, we must first establish clear criteria for what constitutes an optimal layout;
In this work, we define a layout to be optimal if it maximizes space usage (i.e., storage capacity) given a required number of accessible storage points (i.e., pick faces), a required number of longer-term storage points (locations incurring an accessibility penalty) and the physical and functional constraints of the space. 
We aim to design layouts that account for different user preferences among the competing priorities of maximizing storage and accessibility. Therefore, there is generally more than one optimal layout, given the multi-objective nature of the problem.
Hence, we must solve a constrained optimization problem that maximizes storage capacity and number of pick faces while satisfying a set of constraints.
It is worth mentioning that this problem differs from general floor planning, which is a partitioning problem with a given list of rooms and their adjacency, size, and position constraints. 
Thus, the approaches do not quite apply to our problem.

Unlike automatic floor plan generation ~\cite{medjdoub2000separating,che2017bi, wu2018miqp, hu2020graph2plan, laignel2021floor, morisset2021floor}, automatic warehouse layout generation has not been fully explored in the literature. In warehouse environments, there are additional challenges not included in those analyses, such as industrial constraints, changing preferences, and real impact on warehousing operational activities.
Prior attempts to automate warehouse layout design have focused on using mathematical optimization methods~\cite{yener2019optimal}. For example, \cite{zhang2006combining} formulate the problem as an Integer Linear Program (ILP) with the combination of path relinking and a Genetic Algorithm (GA). Gu~\cite{gu2005forward} uses Generalized Benders Decomposition (GBD) to find the optimal solution. Mathematical approaches have some limitations, such as modeling complexity, lack of flexibility in case of any changes in the requirements, and large computational costs.

We propose a new framework to address the gap in the existing literature and to build a tool applicable to real-world scenarios. We present an interactive and iterative tool that allows warehouse designers to impose operational constraints or preferences and evaluate the optimality using objective measures such as capacity and accessibility.  This leads the users to an informed decision on the final configuration based on the existing demands and solves a constrained optimization problem that maximizes storage capacity and number of pick faces while satisfying given warehouse constraints.

\section{Methodology}
We aim to create warehouse layouts that balance various factors such as storage capacity, sufficient number of access points, ease of navigation, and average projected throughput during item retrieval. 
After generating a range of candidate layouts, an experienced warehouse designer can select from the candidates or further refine them. For any candidate chosen in this interactive selection process, the layout would then undergo a thorough validation by an on-site team prior to implementation.

\begin{figure*}[thb]
\centering
		\fbox{\includegraphics[width=0.95\textwidth]{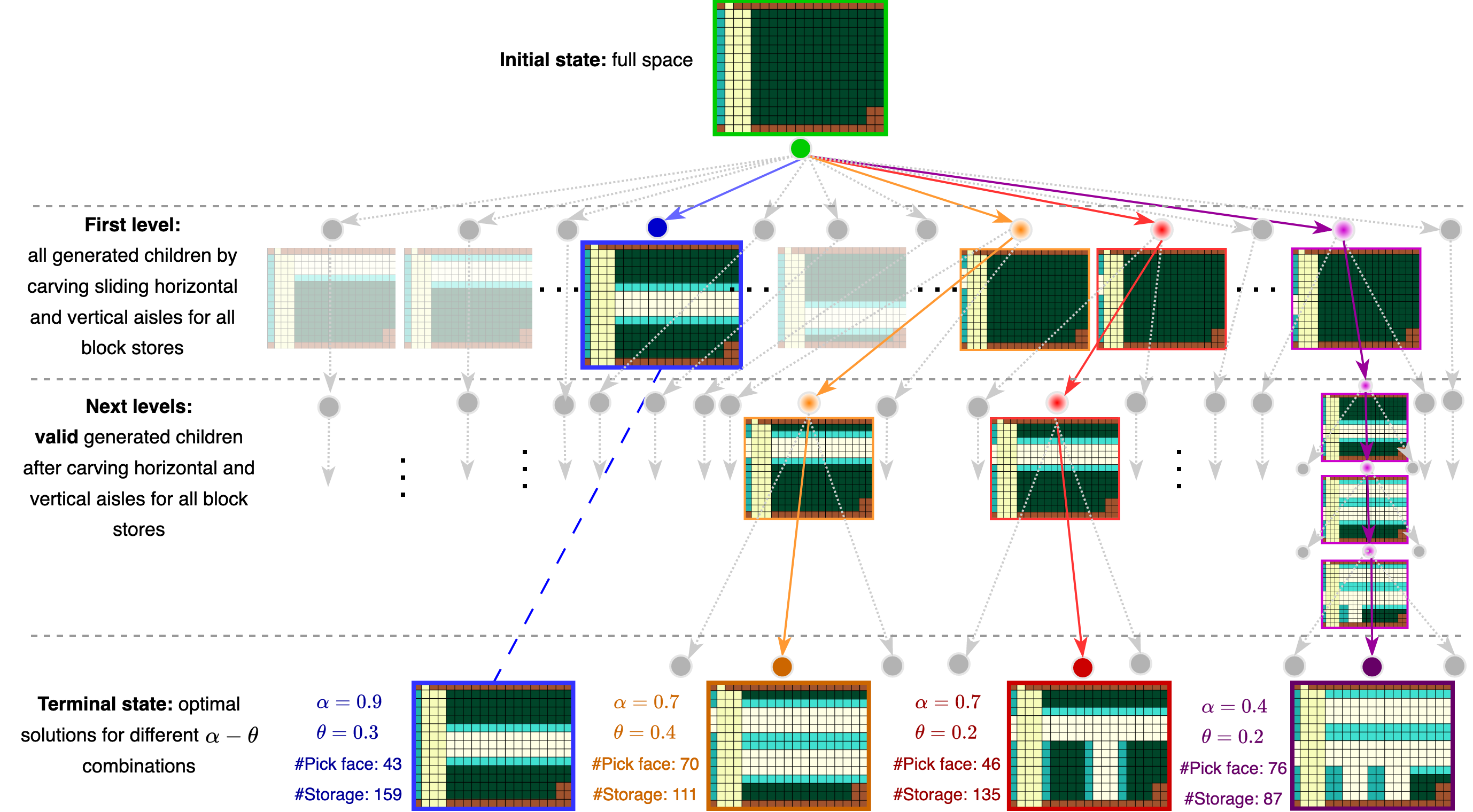}}
		\caption{Running example of beam search ($b=1$) for a sample space from our industry partner with specified door connections and aisle width = 3. At the initial state, all free spaces are assigned as storage spaces. At the first level, all possible unique children are generated by carving aisles with sliding horizontal and vertical aisles across all block stores.  Depending on the particular $\alpha-\theta$ combination specified, a different configuration is found as the best solution for that setting.  This carving process is undertaken again on the chosen configuration in subsequent levels until a terminal state is found.  In the figure, each colored path represents a route that leads to the best possible solution for a particular $\alpha-\theta$ combination.  For each of these solutions, the $\alpha$, $\theta$, number of pick faces and number of storage are shown.
  Space specifications: \legendsquare{RawSienna}~Walls, \legendsquare{door}~Door connections, \legendsquare{aisle}~Aisles, \legendsquare{storage}~Storage, \legendsquare{Turquoise}~Pick face.
  }
		\label{fig:running_exp}
\end{figure*}

We propose a novel candidate layout generation algorithm (see Algorithm~\ref{alg:search}) to generate optimal layouts based on tree search. A given space is specified by a discrete two-dimensional grid of cells with several masks marking the positions of walls $M_{walls}$ and door connections $M_{door\_connections}$. 
Figure~\ref{fig:running_exp} shows a running example of a sample space from our industry partner. Each unit cell in the grid is colored based on what category it belongs to: walls, door connections, aisles, storage, or pick face. 
The grid is initialized with all cells fully occupied $L_{\text{\it{full}}}$. The tree search then explores the space of possible layouts by systematically carving new aisles. Invalid nodes (i.e., layouts violating any constraints) are filtered in the Layout Filtering step (explained in Section \ref{sec:filter}). The valid layouts are scored using a custom scoring function, and those with the highest score are designated as optimal. As discussed before, there are typically multiple optimal (and viable) layouts for consideration by the customer. In Figure~\ref{fig:running_exp}, each colored path represents a route that leads to the best possible solution for a particular setting.

\begin{algorithm}[thb]
  \caption{Candidate Layout Generation (beam size =1)}
  \label{alg:search}
  \begin{algorithmic}[1] 
    \Inputs{$M_{\text{\it{walls}}}, M_{\text{\it{door\_connections}}}$ //input masks} 
    \Initialize{Start from the full space : $L_{\text{\it{full}}}$}
    \State $Q \gets \text{\it{Queue}}()$ //create an empty queue 
    \State $Q.\text{\it{push}}(L_{\text{\it{full}}})$
    \State $L_{\text{\it{optimal}}} \gets L_{\text{\it{full}}}$
    \While{not $Q.\text{\it{empty}}()$}
      \State $L \gets Q.\text{\it{pop}}()$
      \State {Generate all children by carving aisles horizontally and vertically for all block stores in $L$} 
      \If{$L == L_{\text{\it{\text{\it{full}}}}}$} // first level
        \State score valid children
        \If{highest score $> L_{\text{\it{optimal}}}$ score}
            \State $L_{\text{\it{optimal}}}\gets$ valid children with highest score
        \EndIf
        \State $Q.\text{\it{push}}($all children$)$
        \State break
      \EndIf
      \For{All children ($c_1,\ldots,c_n$)}
        \If{$c_i$ is valid}  //filtering step
            \State {Score $c_i$}
        \EndIf
      \EndFor
      \If{children highest score $>= L$ score}
        \State $Q.\text{\it{push}}($children with highest score$)$
        \If{children highest score $> L_{optimal}$ score}
        \State $L_{\text{\it{optimal}}} \gets $children with highest score
        \EndIf
      \EndIf
    \EndWhile \\
    \Return $L_{\text{\it{optimal}}}$
  \end{algorithmic}
\end{algorithm}

\subsection{Tree Search}
Due to the time-intensive nature and memory constraints of exhaustive tree search at scale, we employ beam search for exploring the tree. Beam search is a heuristic Breadth-First Search (BFS) algorithm that helps to make local decisions and limit the search space. In the default setting (beam size: $b=1$), for each node in the tree: (i) all children are generated by carving new possible horizontal or vertical aisles in each block store\footnote{A group of contiguous storage cells is called a block store.}; (ii) child nodes at the first level are always expanded (to promote diversity of solutions). For deeper levels, only valid child nodes are expanded;
and (iii) all valid children are scored and the most promising child (the one with the highest score), is selected for further expansion. The remaining child nodes are pruned. This process continues until there are no more child nodes left for expansion in the tree. 
This indicates that the terminal state has been reached and the layouts with the highest score are taken to be optimal.
For larger beam sizes ($b > 1$) however, $b$ top-scored children are selected at each level. 

\subsection{Layout Filtering}
\label{sec:filter}
To ensure that only viable and efficient layouts are selected, we sift through all generated children layouts and reject those that violate any functional or efficiency constraints as defined below: 

\textbf{Functional constraints:}
\begin{itemize}
    \item Aisles that are connected to pick faces can not be narrower than the specified aisle width, 
    \item All aisles need to be reachable by all doors into the warehouse space,
    \item No item is allowed to be placed in doorways or areas marked as ``reserved,'' and
    \item No pillar can block an aisle.
\end{itemize}

\textbf{Efficiency constraints:}
\begin{itemize}
    \item Aisles wider than the minimum required size are not allowed as they waste space,
    \item Two-sided access block stores should contain at least two rows, and
    \item Each block store should contain more than one item as it is never efficient or desirable to store a single item at a given location. 
\end{itemize}

\subsection{Layout Scoring}
\label{sec:score}
Candidate solutions are evaluated and the underperforming tree nodes are pruned. The scoring function is a critical component used as a heuristic in the tree search. A misspecified score will be detrimental to the node expansion of the tree search resulting in sub-optimal solutions. 
We introduce a scoring function, Eq.~\eqref{equ:fitness}, which not only enables trade-offs between important performance factors (e.g., storage and accessibility) but also facilitates diverse layout generation. We use normalization and define the term weights carefully in the scoring formula to ensure the layout assessment is accurate and unbiased.
\begin{equation}
    \text{\it{Score}} = \alpha T_{s}+\beta T_{pf}+c_1 T_{o},
    \label{equ:fitness}
\end{equation}
The scoring function is a weighted combination of three terms, namely, the normalized storage capacity $T_s$, the normalized number of pick faces $T_{pf}$, and the normalized number of block stores in a specific orientation (vertical or horizontal) $T_o$. The coefficients $\alpha$ and $\beta$ specify the relative importance of the first two terms, while the coefficient $c_1$ is a fixed hyperparameter selected empirically (more details later). 

The normalized storage capacity $T_s$ is defined as,  
\begin{equation}
    T_{s} = \frac{N_s- c_2\theta P_a}{\text{\it{Total open area}}},
    \label{equ:fitness11}
\end{equation}
where $N_s$ is the storage capacity; $\theta$ is a weighting coefficient; $c_1$ and $c_2$ are hyperparameters (discussed in more details later), and $P_a$ is an accessibility penalty defined as,
\begin{equation}
    P_a := \sum_{BS_i} \omega_i \max_{j \in BS_i}(h_j-1)^2,
\end{equation}
with $\omega$, and $h$ corresponding to the width and item heights of a block store $BS$. The penalty is related to the number of items that need to be removed to access the deepest row in a block store. The chosen quadratic scaling has desirable symmetry properties ensuring that higher depths are appropriately penalized.


The second term in Eq.~\eqref{equ:fitness}, $T_{pf}$, the normalized number of pick faces is defined as,

\begin{equation}
    T_{pf} = N_{pf}/N_s,
    \label{equ:fitness12}
\end{equation}

where $N_{pf}$ is the number of pick faces.

Finally, the third term in Eq.~\eqref{equ:fitness}, $T_{o}$ represents the normalized number of block stores in a specific orientation (vertical or horizontal) if that orientation is opposite to the space orientation. Without using this term, block stores tend to be aligned with the space orientation. However, sometimes the opposite orientation is preferable due to the location of the staging area\footnote{Staging area is where items are loaded to or unloaded from the warehouse.}. $T_{o}$ provides a control to allow the dominated layout orientations thereby promoting more diversity in the results.

The weighting coefficients, $\alpha$, $\beta$, and $\theta$, adjust the balance between the different terms. The term $\alpha \in \{0.1,\dotsc, 1\}$ controls the balance between storage capacity and number of pick faces, $\beta = \min(0.1, 1-\alpha)$, and $\theta \in \{0.1,\dotsc, \alpha/2\}$ are defined based on $\alpha$. The term $\theta$ controls the accessibility penalty $P_a$. The maximum of $\theta$ is set based on $\alpha$ to avoid over-penalizing. Hyperparameters $c_1=0.01$ and $c_2=0.1$ are set empirically to scale the corresponding terms. 


Each combination of $\alpha$ and $\theta$ represents a particular preference for the properties of the generated optimal layout discovered by the tree search. The process for selecting the final layout involves the warehouse designer, and is discussed in more details in Section~\ref{sec:res}.

\subsection{Connectivity Score}
In addition to the scoring function presented above, we define another score term, connectivity to estimate the likely relative throughput expected from different candidate layouts. Note that the exact throughput cannot be known \textit{a priori} as it will depend on specific product assignment and order lists. Our estimates here are used only to determine the ranking order of different candidates as a tool to select among several optimal candidates generated above. The connectivity score is defined as the average cumulative ratios of shortest distances to the Manhattan distances~\cite{black2006manhattan} for pairs of pick faces:

\begin{equation}
    \mathcal{C} = \frac{1}{N_{pf}} \sum_{i \leq j} \frac{1}{D_{i,j}^{\text{\it{Shortest}}} / D_{i,j}^{\text{\it{Manhattan}}}},
\end{equation}
where $N_{pf}$ is the number of pick faces, and
$D_{i,j}$ is the distance (i.e., shortest or Manhattan) between two pick face locations. 

This is based on the intuition that for a high-throughput (i.e., more connected) layout, the shortest distances (between pairs of pick faces) will on average be closer in value to the Manhattan distance. This function has the additional desired property that it is normalized to one, facilitating simple direct comparisons between different layouts. Note that we decided not to include connectivity as an independent scoring term in Eq.~\eqref{equ:fitness} to simplify the process and minimize instabilities in the search.

\begin{figure*}[t]
\centering
		\fbox{\includegraphics[width=0.95\textwidth,trim=0 0 0 2,clip]{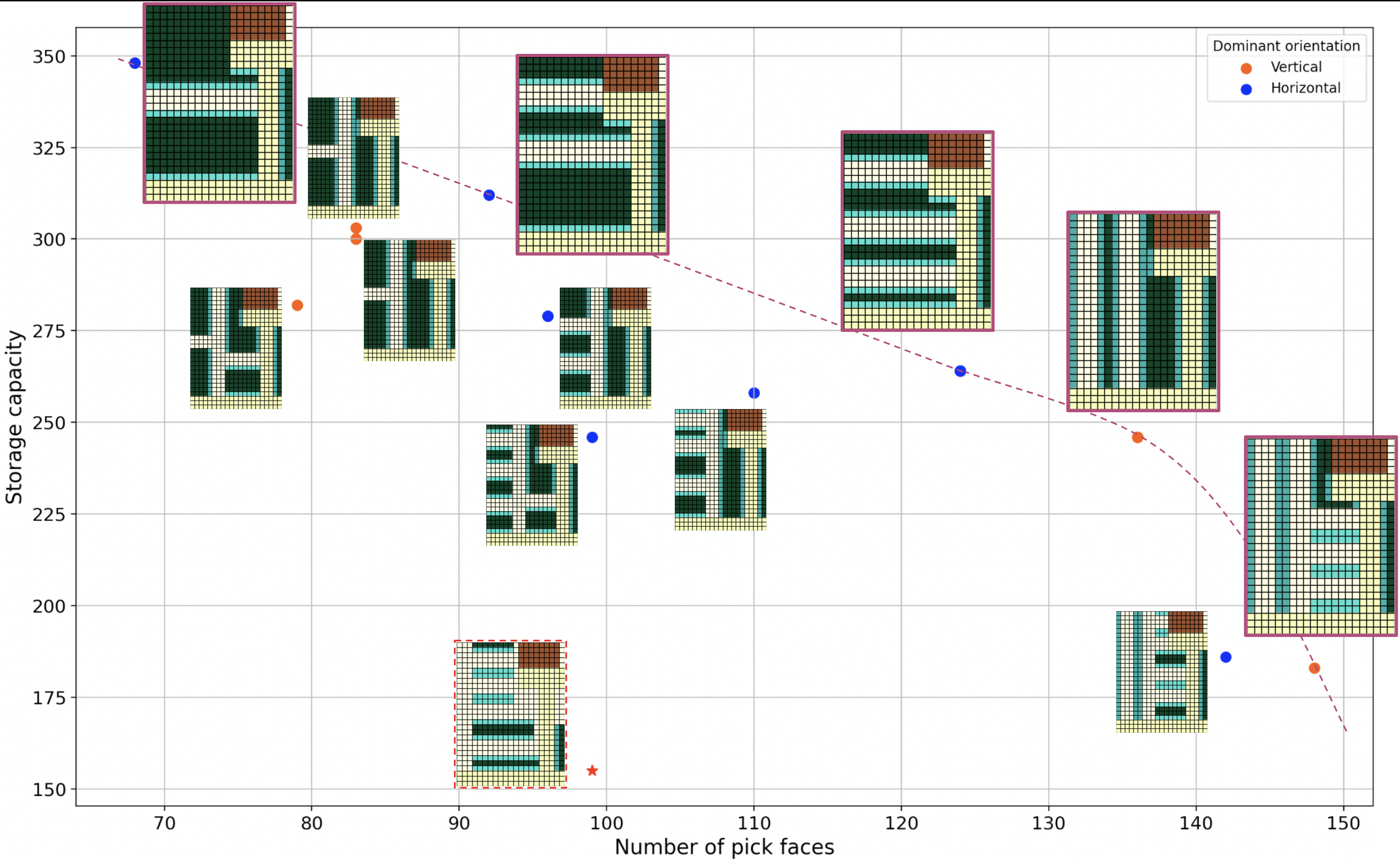}}
		\caption{Pareto visualization for a medium-sized space from our industry partner with specified door connections and aisle width = 3. The pink dashed line shows the Pareto front. Zoomed layouts correspond to the data points on the Pareto front. The red star at the bottom shows the manually-designed layout which has been specified by the red dashed border. 
  Space specifications include the following: \legendsquare{RawSienna}~Walls, \legendsquare{door}~Door connections, \legendsquare{aisle}~Aisles, \legendsquare{storage}~Storage, \legendsquare{Turquoise}~Pick face.
  }
		\label{fig:pareto}
\end{figure*}
\subsection{Post-Refinement}
\label{sec:post}
The objective of the post-refinement step is to apply any additional constraints to arrive at the final layouts. While every application will require some customization by the on-site team often these constraints and requirements can be codified to save time. In one application, the pallet racking system only allowed even numbers of racking units along the total block store (due to how the racking infrastructure was constructed).  In another application, clear paths were necessary to access pillars that contained fire safety equipment. 
In both cases, the flexibility of our algorithm allowed for these constraints to be programmatically applied as a final step, only passing layouts that fulfilled the criteria.

\subsection{Implementation Details}
Our method is adjustable to larger beam sizes. By increasing the beam size, the search space is expanded as more children are explored at a time.
However, it also increases the exploration time. Therefore, we use multiprocessing to expedite the search process, especially for larger beam sizes. We used eight CPU cores with 16GB of memory to run experiments. For a medium-sized space ($50\times55$), the average processing time for generating the Pareto plot (e.g., Figure~\ref{fig:pareto}) is 85s.

\section{Results}
\label{sec:res}
To generate all possible optimal layouts for a given space, the layout generation process is run separately for all combinations of $\alpha$ and $\theta$ in their defined ranges. 
Drawing from the pool of generated optimal layouts, we create a Pareto plot that visualizes possibilities with respect to the storage capacity and number of pick faces. The Pareto plot is used as an interactive decision-making tool.
Figure~\ref{fig:pareto} shows the Pareto plot for a medium-sized space from our industry partner.
The Pareto front comprises optimal layouts that dominate the other candidates by striking a better trade-off between the storage capacity and number of pick faces.
The Pareto front is downward sloping, illustrating that the storage capacity decreases with increasing number of pick faces.
When two candidate layouts have the same score in the same $\alpha-\theta$ setting, the connectivity score comes into play and decides which one is likely to lead to a higher-throughput design.

In Figure~\ref{fig:pareto}, we compare the auto-generated optimal layouts in the Pareto plot with the existing manually-designed layout (indicated by the red star added to the plot). We observe notable improvements across both number of pick faces and storage capacity.


We closely collaborated with expert warehouse designers to validate the quality of the generated layouts and compare the results with manually-generated versions. The designers reported that the process was user-friendly and served as an effective collaborative tool, significantly streamlining their efforts to achieve good layouts.


\section{Discussion}
Warehouse layout design plays a vital role in warehouse operations performance. We proposed a novel automated optimal layout candidate generation framework using beam search that satisfies a set of constraints. We also introduced a new scoring function that handles a balance between storage capacity, number of access points, and accessibility cost. Our method can generate a wide variety of candidate layouts for different ranges of picking and storage areas and we demonstrated this in various spaces of two real-world warehouses. The simplicity of the method makes it easily adaptable to any changes in user specifications and requirements. 

Despite all these strengths, our approach is not without limitations. One limitation is that we were unable to measure the throughput of the layouts and compare their performances comprehensively. Throughput depends on order lists and item allocation, which would add a layer of complexity that is beyond the scope of this work.  As throughput is the ultimate measure of effectiveness, not being able to account for these factors restricts our ability to fully evaluate the efficiency of the different layouts. 
Despite this limitation, the proposed solution was tested on 9 available physical spaces through the partnership between Amii and Routeique. In the experiments, the framework generalized well to both large and small spaces as well as those with non-rectangular shapes and the expert warehouse designers verified the feasibility and quality of the generated layouts.

Finally, we did not incorporate some relevant constraints, for example varying heights of rows, and larger-sized doorways/entryways. However, we found that it was relatively straightforward to programmatically add those additional constraints for the unique artifacts found in spaces such as fire extinguishers (see Section~\ref{sec:post}). While it is challenging to comprehensively anticipate and account for all unique elements found in specific real-world scenarios, our method is flexible and could incorporate the majority of such constraints.

In the future, we will continue to validate and refine our tool in more diverse settings and in different warehouses. Feedback from a larger number of warehouse designers will guide the direction of future development.

\section*{Acknowledgments}
We thank Mara Cairo, Bevin Eldaphonse, and the Routeique team, especially Mike Allan for their valuable support during this project. We also thank our reviewers for their constructive feedback, which helped us to improve the quality of this work.

\bibliographystyle{unsrtnat}
\bibliography{references}  






\end{document}